\documentclass[runningheads]{llncs}
\usepackage[T1]{fontenc}
\usepackage{hyperref}
\usepackage{url}
\usepackage{subfigure}
\usepackage{multirow}
\usepackage[table]{xcolor}
\usepackage{appendix} 
\usepackage{listings}
\usepackage{xcolor}
\usepackage{enumitem}

\usepackage{graphicx}
\begin{document}
\title{LERO: LLM-driven Evolutionary framework with Hybrid Rewards and Enhanced Observation for Multi-Agent Reinforcement Learning}
\titlerunning{LERO}
%
\author{Yuan Wei \and Xiaohan Shan \and Jianmin Li}
\authorrunning{Y. Wei et al.}
%
\institute{Qiyuan Lab, Beijing, China \\
\email{\{weiyuan,shanxiaohan,lijianmin\}@qiyuanlab.com}}
\maketitle              
\begin{abstract}
Multi-agent reinforcement learning (MARL) faces two critical bottlenecks distinct from single-agent RL: credit assignment in cooperative tasks and partial observability of environmental states. We propose LERO, a framework integrating Large language models (LLMs) with evolutionary optimization to address these MARL-specific challenges. The solution centers on two LLM-generated components: a hybrid reward function that dynamically allocates individual credit through reward decomposition, and an observation enhancement function that augments partial observations with inferred environmental context. An evolutionary algorithm optimizes these components through iterative MARL training cycles, where top-performing candidates guide subsequent LLM generations. Evaluations in Multi-Agent Particle Environments (MPE) demonstrate LERO's superiority over baseline methods, with improved task performance and training efficiency.

\keywords{Multi-Agent Reinforcement Learning \and Hybrid Rewards \and LLM \and Evolutionary Algorithms \and observation enhancement.}
\end{abstract}
\section{Introduction}
Reinforcement learning has achieved remarkable success in solving decision-making problems. 
As many real-world scenarios involve multiple interacting agents, MARL has emerged as a powerful paradigm for solving complex tasks.
Among various MARL frameworks, centralized training with decentralized execution (CTDE) has become the dominant approach, as it effectively balances the benefits of centralized information during training with the practical constraints of decentralized deployment in applications such as autonomous driving, robotics teamwork, and multi-player games.
CTDE-based MARL algorithms face two key challenges: credit assignment and partial observability.
Credit assignment involves determining individual agents' contributions to team success, while partial observability forces agents to make decisions with limited environmental information. 
Both factors significantly complicate the learning process and limit performance.

Recent advances in LLMs have demonstrated remarkable capabilities in understanding complex scenarios, reasoning about causal relationships, and generating contextual information. 
These capabilities make LLMs potential experts that can guide MARL training by designing algorithm modules that address the aforementioned challenges through their contextual understanding and reasoning abilities.

In this paper, we present an LLM-driven Evolutionary framework with hybrid Rewards and enhanced Observation (LERO), a novel approach that differs from previous work by combining the reasoning power of LLMs with the optimization capabilities of evolutionary algorithms.
Our approach employs an evolutionary algorithm at the outer loop to optimize two key components generated by LLMs: hybrid reward functions (HRFs) that combine global team objectives with individual agent contributions, and observation enhancement functions (OEFs) that complement missing global state information.

Our main contributions are as follows.
\begin{enumerate}[label=\arabic*)]
\item Novel LLM-based hybrid reward functions and observation enhancement functions that effectively address the credit assignment and partial observability challenges in MARL;
\item A novel evolutionary framework that systematically optimizes LLM-generated components through iterative feedback, creating a closed-loop improvement process for MARL training;
\item Comprehensive experimental results in MPE scenarios demonstrating how our evolutionary framework effectively improves the quality of LLM-generated components and enhances overall MARL performance.
\end{enumerate}

The remainder of the paper is organized as follows. 
Section \ref{Related work} reviews related work in the field, highlighting key advancements and existing challenges. 
Section \ref{Method} presents the methodology, detailing the framework design and the algorithmic workflow. 
Section \ref{Experiments} describes the experimental setup and evaluation metrics, followed by a discussion of the experimental results. 
Finally, Section \ref{Conclusion} concludes the paper, offering insights and potential directions for future research.

\section{Related work}
\label{Related work}
Prior research has explored two critical dimensions of reinforcement learning systems: methods for generating effective reward functions that properly assign credit among agents, and techniques for enhancing observation quality to overcome partial observability limitations. These two dimensions directly address the key bottlenecks in MARL that we identified in the introduction.
\subsection{Reward generation}
Designing rewards in MARL presents significant challenges, including credit assignment, ensuring coordination among agents, and dealing with non-stationarity in learning environments.
Several approaches have focused on how to effectively combine individual and global rewards for agents, such as Matsunami \cite{matsunami2021reward}, who integrated individual behavior rewards with penalties reflecting negative contributions to social welfare, leading to increased social utility among agents and improved cooperative policy formation.  
Wang \cite{wang2022individual} employed the Individual Reward Assisted Team Policy Learning method, which combines individual and team policies with discrepancy constraints to enhance cooperation in MARL scenarios. 
However, these approaches still rely on manually designed reward functions for specific scenarios, limiting their generalizability across different multi-agent tasks.

With the powerful understanding and reasoning capabilities of LLMs, an increasing number of researchers are leveraging them to design reward function. 
A significant portion of this work has focused on single-agent reinforcement learning algorithms\cite{cao2024survey}.
Ma \cite{ma2024eureka} and Xie \cite{xie2024text2reward} leverage large language models to automatically generate executable, interpretable dense reward functions from natural language goal descriptions, producing superior reinforcement learning outcomes that consistently outperform human expertise across diverse robotic tasks.
Kim \cite{kim2023guide} and Wang \cite{wang2024rlvlmf} expand research in the multimodal domain.
However, there is still a significant research gap in utilizing large language models to assist reward function design in multi-agent reinforcement learning, particularly for addressing the credit assignment problem where individual contributions to team success must be accurately evaluated.

\subsection{Observation enhancement}
Partial observability presents a fundamental challenge in MARL, where agents must make decisions with limited environmental information. 
The foundational work in observation enhancement began with Hausknecht and Stone \cite{hausknecht2015deep} introducing the Deep Recurrent Q-Network (DRQN).
By incorporating recurrent neural networks, they enabled agents to leverage historical information to enhance current observations. 
Graves \cite{graves2016hybrid} introduced Memory-Augmented Neural Networks, which utilized external memory modules for storing and retrieving historical information, significantly improving agent performance in partially observable environments.
Liu \cite{liu2020multi} proposed a graph attention network-based approach that employs dynamic weighting mechanisms for information filtering. 
Sheng \cite{sheng2022learning} developed a hierarchical attention architecture that separately processes temporal and spatial information dependencies.
However, these methods primarily focus on the structural aspects of information processing while lacking a semantic understanding of the environment.

Recent works of LLMs have demonstrated promising applications in observation enhancement. 
Paischer \cite{paischer2024semantic} pioneered using language models to compress agent histories into semantic descriptions and developed human-readable memory mechanisms for RL, significantly improving performance on long-term dependency tasks. 
Meanwhile, Radford\cite{radford2021learning} showed how natural language supervision can enable transferable visual representations through their CLIP model, establishing the effectiveness of cross-modal learning. 
While these studies focused primarily on single-agent environments, the potential of LLM-enhanced observation in multi-agent settings remains largely unexplored, particularly for dynamically inferring global state information from partial observations to facilitate more effective team coordination.

Building upon these foundations in reward generation and observation enhancement, our work introduces a novel evolutionary framework that leverages LLMs to address both challenges simultaneously in MARL settings.
Unlike previous approaches that focus on either reward design or observation enhancement in isolation, LERO integrates both aspects within an evolutionary optimization process, creating a comprehensive solution specifically tailored to the unique demands of multi-agent systems.

\section{Method}
\label{Method}
\begin{figure}[htbp] 
    \centering 
    \includegraphics[width=0.90\textwidth]{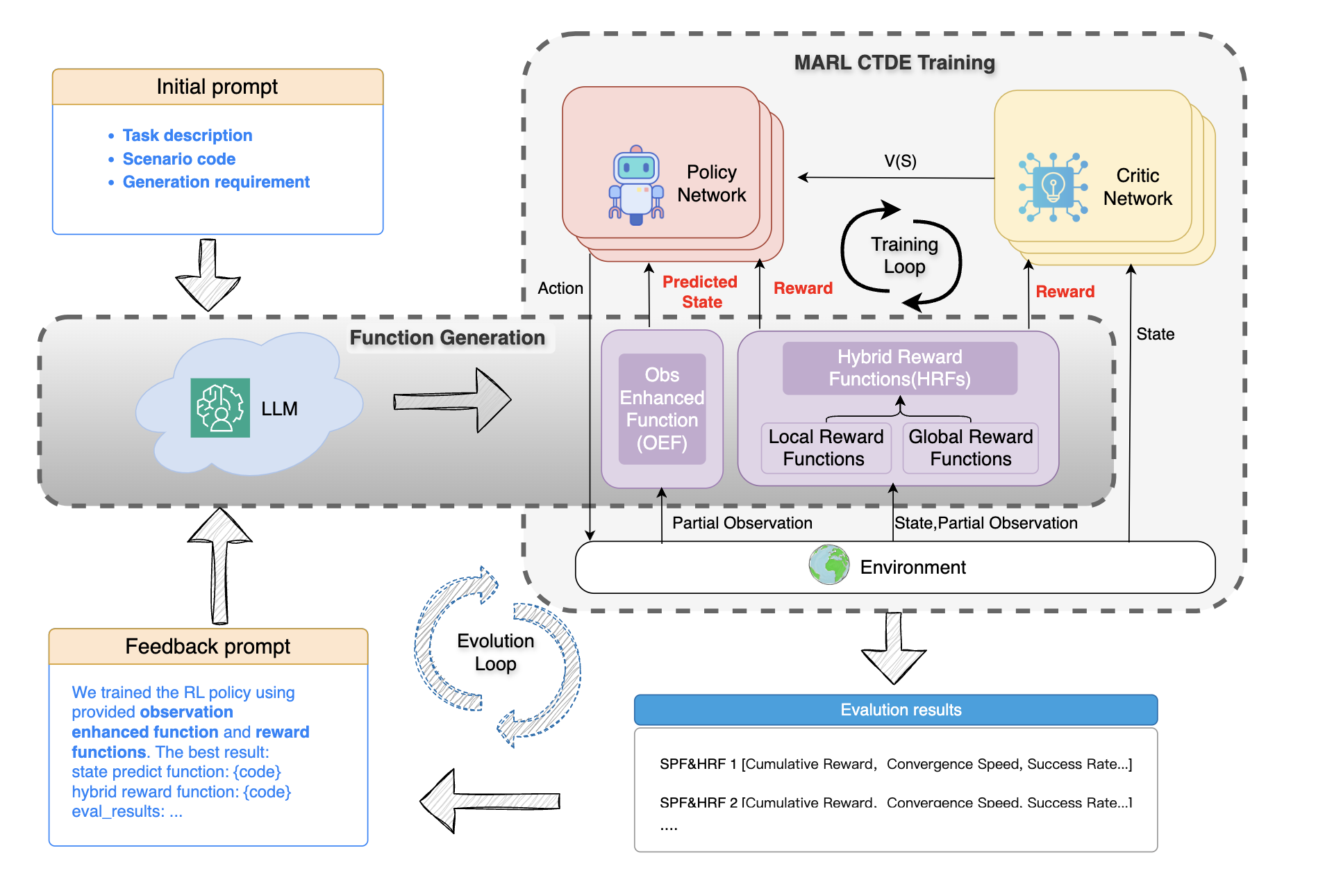} 
    \caption{The LERO framework follows an iterative process where HRFs and OEF are generated by LLMs based on task descriptions, environment code, and evolution descriptions. In each iteration, a selector module evaluates the performance of these HRFs and OEF, allowing for the selection of the most effective  components for MARL training, ultimately enhancing agent adaptability and cooperation.} 
    \label{fig:framework} 
\end{figure}

In this section, we introduce LERO, a framework combining LLMs with evolutionary algorithms to address the twin challenges of credit assignment and partial observability in MARL (Fig.~\ref{fig:framework}). 
We first provide an overview of the framework architecture, then detail each key component: the LLM-based generation of hybrid reward functions (HRFs) and observation enhancement functions (OEFs), the evolutionary optimization process, and the integration with MARL training.
The framework leverages evolutionary principles: selection (choosing the best-performing candidates), crossover (combining features), and mutation (introducing variations) to optimize solutions through an iterative process inspired by natural selection.

In our framework, LLMs perform the role of evolutionary operators such as crossover and mutation. 
Specifically, the LLM generates new Hybrid Reward Functions (HRFs) and Observation Enhancement Functions (OEFs) by combining components of high-performing candidates (crossover) and introducing novel variations or adjustments to their structure (mutation). 
A Selector module guides these evolutionary operations by evaluating the performance of these functions in a MARL environment.

\subsection{Reward Function and Observation Enhanced Function Generation}
Our framework initializes by feeding task descriptions and environment specifications into LLMs to generate task-specific HRFs and OEF.

The resulting HRFs synthesize individual reward signals, which assess agent-specific behaviors, with global reward metrics that evaluate collective team performance. Mathematically, we can express this as: \begin{equation}
R_i(o_i, S, a_i) = \alpha_i · R_{local}(o_i, S, a_i) + (1-\alpha_i) · R_{global}(S, a)
\end{equation}
where $R_{local}(o_i, S, a_i)$ evaluates individual contributions, $R_{global}(S, a)$ assesses team performance, and $\alpha_i$ balances these components based on the specific task requirements.

Additionally, LLMs generate OEF that transform the original observation space into an enhanced observation space: 
\begin{equation}
o'_i = OE(o_i, h_i)
\end{equation}
where $o'_i$ represents the enhanced observation for agent $i$, and $h_i$ denotes the agent's observation history. 
These enhanced observations provide richer contextual information to support policy agents' decision-making processes.
In our notation, S represents the current state, $o_i$ denotes the observation of agent $i$, $a_i$ is the action taken by agent $i$, and $a = {a_1, ..., a_n}$ signifies the set of actions performed by all agents.
To maintain clarity, we have omitted the time step $t$ when it does not introduce any ambiguity.

Each iteration involves parallel LLM inference to generate multiple candidate HRFs and OEFs for MARL training. 
Importantly, the LLM inference occurs only during the function generation phase and not during the actual MARL training, which significantly reduces computational costs.

\subsection{MARL Training}
In the MARL training phase, the framework utilizes various reward functions and observation enhancement functions generated by the LLMs in parallel training instances. 
Each training instance implements a specific HRFs and OEF combination, serving as an evaluator for their effectiveness.

To enhance the efficiency of this evaluation process, we reduce the number of training iterations needed for full convergence, allowing us to obtain a preliminary assessment of each function combination's performance. 
We evaluate these functions using metrics such as cumulative team reward, convergence speed, and task-specific success rates.

This stage-wise evaluation provides valuable performance data, which serves as input for the Selector module. 
The LERO framework is designed to be algorithm-agnostic, accommodating different MARL algorithms that can be selected based on the specific task or environment requirements.

\subsection{Selector Module}
The Selector module uses specific criteria to rank the evaluation results of each HRFs and OEF combination and selects the most effective ones, providing feedback to the LLMs for the next generation.
One simple selection criterion is to select the top $k$ performing combinations based on predefined metrics such as cumulative reward, convergence speed, or success rate. 
Additionally, other selection strategies from evolutionary algorithms can be incorporated. 
For instance, techniques such as tournament selection and rank-based selection can be adapted to choose functions based on their relative performance, which fosters a competitive environment among the candidates.
These strategies have been well-established in evolutionary computation literature, including studies by Goldberg \cite{Goldberg1988GeneticAI} on genetic algorithms, Deb \cite{deb2001multiobjective} on multi-objective optimization, and Bäck et al. \cite{back2023evolutionary} regarding function optimization.

By allowing for customizable selection rules, the LERO framework can effectively adapt to different scenarios, improving the overall performance and collaboration of agents in MARL tasks.

\subsection{Feedback Loop}
The entire process operates through an iterative feedback loop that leverages performance feedback from the Selector to refine HRFs and OEF designs. 
This mechanism simulates the crossover and mutation processes in evolutionary algorithms, enabling LLMs to generate new functions in each iteration.

Specifically, the LLM receives the following inputs for each new generation:
\begin{enumerate}[label=\arabic*)]
  \item The original task and environment descriptions.
  \item Code of the best performing HRFs and OEF from previous iterations.
  \item Performance metrics and analysis of these functions.
\end{enumerate}

These newly generated functions are developed based on historical performance analysis and Selector feedback, aiming to enhance overall cooperation through more precise agent behavior evaluations and improved environmental understanding.
By integrating these steps, the LERO framework effectively combines LLMs' generative capabilities with evolutionary algorithm principles, providing an innovative solution to both credit assignment and partial observability challenges in multi-agent environments.

\section{Experiments}
\label{Experiments}
\subsection{Experiment Setup}

\subsubsection{Environments.}
We validated our approach using two cooperative navigation tasks from the Multi-Agent Particle Environment (MPE) \cite{lowe2017multi}: simple spread and simple reference. 
In these scenarios, $N$ agents need to interact with $M$ landmarks. 
A threshold determines successful landmark detection: when the distance between an agent and a landmark is less than a coverage radius of 0.1, the agent is considered to have successfully monitored that landmark.
The main objective is to maximize unique landmark coverage within a limited time frame, requiring effective collaboration among agents.

Importantly, each landmark can contribute to the reward only once at each timestep. 
This means if multiple agents monitor the same landmark simultaneously, the system registers only one successful detection, making redundant coverage inefficient. 
This constraint encourages agents to distribute themselves optimally across different landmarks to maximize overall coverage efficiency.

\subsubsection{Baselines.}
To validate the effectiveness of the LERO framework, we designed a comprehensive set of comparative experiments.
First, we compared LERO against the native reward functions and observation vectors provided in the standard MPE scenarios as our primary baseline.

Second, we conducted three sets of ablation studies to isolate the contribution of each component:
\begin{enumerate}[label=\arabic*)]
  \item LERO with LLM-generated reward functions only (native observations)
  \item LERO with LLM-generated observation enhancement functions only (native rewards)
  \item LERO with LLM-generated components but without the evolutionary optimization
\end{enumerate}
These ablation experiments allowed us to quantify the individual contributions of each component and identify potential synergistic effects when they operate together within the complete LERO framework.

To ensure algorithm-agnostic results, we implemented our framework with three widely-used MARL algorithms: MAPPO \cite{Yu2021mappo}, VDN \cite{sunehag2017value}, and QMIX \cite{rashid2020monotonic}. 
This diverse algorithm selection demonstrates that LERO's benefits are not limited to specific algorithmic approaches.

Key experimental parameters, including Environment Parameters, Training Settings, and LLM Configurations, are detailed in Appendix \ref{appendix A}. 
The prompt templates used for generating both reward functions and observation enhancement functions are provided in Appendix \ref{appendix B}, demonstrating our structured approach to LLM-assisted MARL enhancement.
The following sections present our experimental results in detail, comparing LERO against baseline approaches and analyzing the contribution of each component through ablation studies. 

\subsection{Experiment Results}
\begin{figure}[htbp]
    \centering
    \subfigure[simple spread]{
        \includegraphics[width=0.45\textwidth]{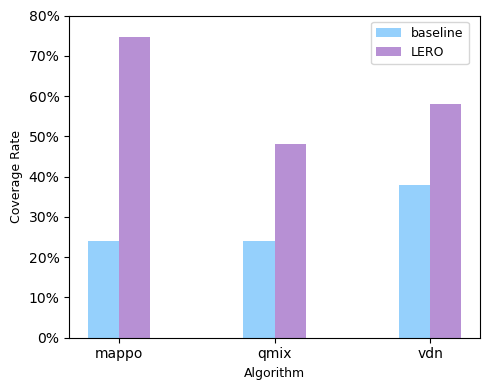}
    }
    \subfigure[simple reference]{
        \includegraphics[width=0.45\textwidth]{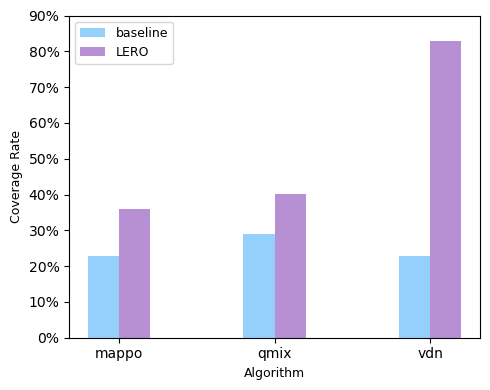}
    }
    \caption{Comparison between LERO Framework and Baseline}
    \label{fig:LERO vs baseline} 
\end{figure}

As shown in Fig.\ref{fig:LERO vs baseline}, the LERO framework consistently outperforms the original MPE implementations across all tested algorithms and environments. The performance improvements are particularly striking in the coverage rates, which measure the percentage of landmarks successfully monitored by agents.

In the simple spread environment, LERO demonstrates substantial gains across all three algorithms. The most dramatic improvement appears with MAPPO, where the coverage rate increases from 0.24 to 0.747—a 211\% improvement. VDN and QMIX also show significant enhancements, with coverage rates improving by 53\% and 100\% respectively. 
These results indicate that the semantically rich reward and observation representations generated by our framework enable much more effective coordination behaviors.

The simple reference task, which requires more complex communication and coordination, also shows marked improvements under the LERO framework. Most notably, VDN achieves an exceptional 261\% increase in coverage rate, from 0.23 to 0.83. QMIX and MAPPO demonstrate more modest but still significant improvements of 38\% and 57\% respectively.
The higher complexity of this task demonstrates LERO's ability to adapt its generated functions to specific environmental challenges.

These results validate that evolutionarily refined reward functions and observation enhancements effectively encode task-specific semantic information that guides agents toward more efficient collaborative behaviors.
More detailed experimental results can be found in App. \ref{appendix C}.

\subsection{Analysis of Function Generation Module}
\begin{figure}[htbp]
    \centering
    \subfigure[simple spread]{
        \includegraphics[width=0.45\textwidth]{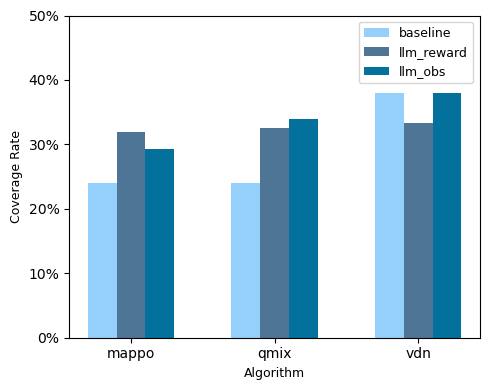}
    }
    \subfigure[simple reference]{
        \includegraphics[width=0.45\textwidth]{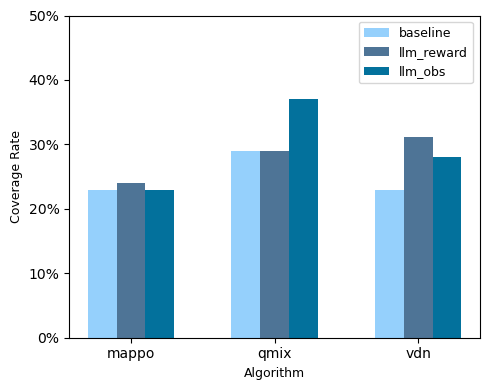}
    }
    \caption{Results of Hybrid-Reward-only and Observation-Enhanced-only Variants}
    \label{fig:reward only} 
\end{figure}

To evaluate the individual contribution of each component, we compared the hybrid-reward-only variant (LR) and observation-enhanced-only variant (LO) separately against the baseline model, using the non-evolutionary versions of each enhancement. 
Fig.~\ref{fig:reward only} demonstrates that both LLM-generated components contributed positively in most scenarios, though with algorithm-specific variations.

In the Spread environment, both the LLM-generated hybrid reward (LR) and observation enhancement (LO) components generally improved performance, with MAPPO showing a 33.3\% improvement with LR and 22.1\% with LO, while QMIX benefited more substantially from LO (41.7\% increase). 
Interestingly, VDN showed sensitivity to the reward modification, maintaining performance with LO but declining with LR. 
This suggests that VDN's credit assignment mechanism may conflict with certain aspects of the LLM-generated reward structure in this environment.
The Reference environment revealed even more algorithm-specific effects, with QMIX showing no improvement from LR but a substantial 27.9\% gain from LO, while VDN exhibited the opposite pattern with a stronger benefit from reward enhancement (35.2\% improvement).

These results highlight the broad effectiveness of both components across our experimental framework, with each component providing measurable improvements in the majority of tested scenarios even without evolutionary optimization.
While some algorithm-specific variations exist, the overall trend demonstrates that both hybrid rewards and observation enhancements contribute meaningful performance gains across diverse MARL settings. 
The data shows that in 5 out of 6 algorithm-environment combinations, at least one component improved performance, with several cases showing substantial gains exceeding 25\%.

To further investigate how large language models inject domain expertise into reinforcement learning systems, we analyze the LLM-generated functions and identify key improvements in both reward formulation and observation processing that contribute to performance gains.

The LLM-generated reward function (App. \ref{appendix D}, Lis.\ref{reward_no_evolutionary}) significantly enhances the original implementation by introducing a dual reward structure. Unlike the baseline which only uses negative Euclidean distance, the LLM version combines continuous distance feedback with substantial bonuses (+10.0) when agents reach proximity thresholds (< 0.1 units). This creates stronger learning gradients while computational optimizations and improved reward aggregation maintain robust signals across multiple agents, demonstrating how LLMs can effectively inject domain expertise into reinforcement learning systems without specialized human intervention.

The LLM-generated observation enhancement function (App. \ref{appendix D}, Lis.\ref{obs_no_evolutionary}) transforms raw agent observations into a more informative geometric representation through two key computations: precise Euclidean distances to each landmark and an estimated distance to the partner agent derived from relative landmark positions. This spatial feature extraction provides agents with explicit distance metrics rather than requiring them to implicitly learn these relationships from position vectors alone. By capturing both environmental (landmark) distances and social (inter-agent) proximity in a compact four-dimensional feature vector, the function supplies high-level spatial awareness that directly supports decision-making in coordination tasks. 

\subsection{Analysis of Evolution Module}
\begin{figure}[htbp]
    \centering
    \subfigure[MAPPO]{
        \includegraphics[width=0.3\textwidth]{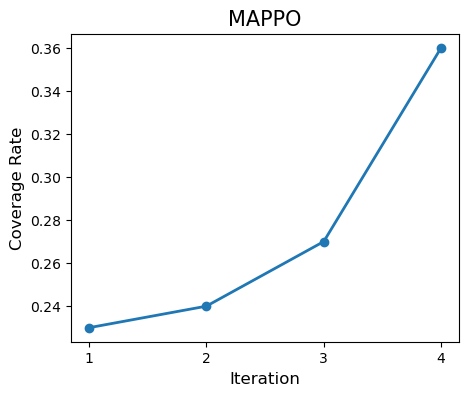}
    }
    \subfigure[QMIX]{
        \includegraphics[width=0.307\textwidth]{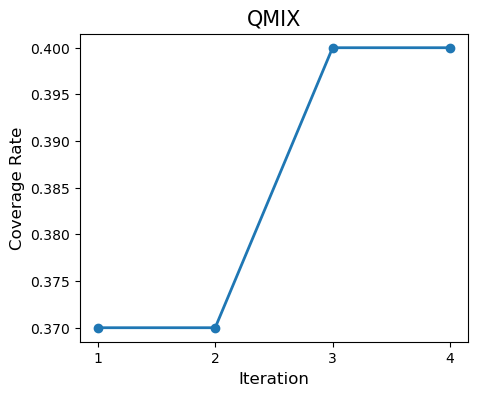}
    }
    \subfigure[VDN]{
        \includegraphics[width=0.3\textwidth]{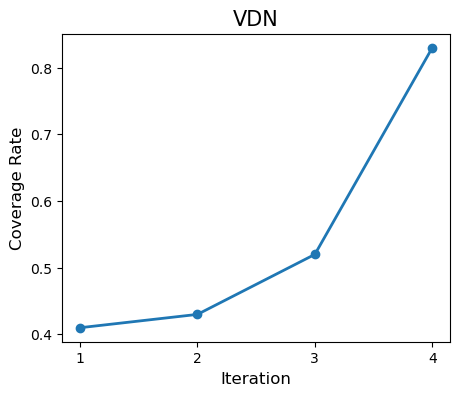}
    }
    \caption{Coverage rate of each iteration in Simple Reference}
    \label{fig:Coverage of iteration} 
\end{figure}

Furthermore, to validate the advantages of our evolutionary design, we recorded the coverage rate for different algorithms in the Simple Reference scenario across four evolutionary iterations. 
Each iteration represents a generation of LLM-refined functions based on feedback from previous performance. 
The results in Fig.\ref{fig:Coverage of iteration} clearly demonstrate that as the evolutionary process progresses, the generated reward functions and state enhancement functions improve the algorithm training outcomes. 

Performance curves show consistent upward trends for all three algorithms in the evolutionary optimization process. 
MAPPO (Fig.~\ref{fig:Coverage of iteration}a) shows modest but steady improvement, with a coverage rate increasing from 0.23 in iteration 1 to 0.26 in iteration 4, representing a 13\% improvement.
QMIX (Fig.~\ref{fig:Coverage of iteration}b) demonstrates more substantial progress from 0.29 to 0.35 (21\% improvement), while VDN (Fig.~\ref{fig:Coverage of iteration}c) exhibits the most dramatic gain, more than doubling from 0.41 to 0.83 (102\% improvement) in iterations. 

This progressive improvement confirms the effectiveness of our evolutionary approach in adapting and refining both reward structures and state representations to better suit the specific requirements of multi-agent coordination tasks.
The significant variation in improvement rates across algorithms suggests that the evolutionary process may be particularly effective in addressing algorithm-specific limitations, with VDN benefiting most substantially from the refined functions.

Furthermore, we analyze the evolutionarily refined reward function and observation enhancement module to understand how these adaptations improve coordination efficiency and learning dynamics.
The LLM-generated reward function with evolutionary (App. \ref{appendix D}, Lis.\ref{reward_evolutionary}) represents a comprehensive upgrade through several key optimizations. 
\begin{enumerate}[label=\arabic*)]
  \item Graduated reward shaping: Introduces a tiered structure providing varying bonuses (15, 8, 4) at multiple distance thresholds (0.1, 0.2, 0.3), creating a more informative learning gradient than the original single-threshold approach.
  \item Proximity incentives: Rewards agents for maintaining appropriate distances to their partners while penalizing excessive separation, encouraging effective positioning for communication.
  \item Stability rewards: Reinforces controlled positioning when agents are near targets, rewarding low velocity at critical moments.
  \item Team synchronization: Includes a substantial collective bonus when all agent pairs simultaneously reach their targets, promoting team synchronization rather than merely individual performance. 
  \item Amplified signal strength: Base penalties and success bonuses have been amplified to create stronger learning signals that better differentiate between effective and ineffective behaviors. 
\end{enumerate}
These evolutionary refinements transform the reward from a simple distance metric into a sophisticated multi-objective signal that encodes specific collaborative behaviors, stability considerations, and graduated progress indicators.

The LLM-generated observation enhancement function with evolutionary (App. \ref{appendix D}, Lis.\ref{obs_evolutionary}) represents a significant advancement over the previous version by extracting task-specific semantic information rather than just generic spatial metrics.
The evolved function introduces several specialized features that directly align with the coordination objectives:
\begin{enumerate}[label=\arabic*)]
  \item Goal-aware perception: Intelligently identifies each agent's assigned target landmark by matching goal colors, allowing for computation of goal-relevant metrics like assigned landmark distance and angle. This task-aware approach means agents can distinguish their own goals from other landmarks. 
  \item Success threshold awareness: Introduces coverage margins that explicitly encode proximity thresholds relevant to task success, providing direct feedback on goal achievement status.
  \item Relational coordination: Captures information between agents' goal progress with features like relative assigned distance difference, enabling agents to reason about comparative performance and adjust strategies accordingly.
  \item Dynamic behavioral information: Extracts movement patterns through velocity norms and communication channel magnitude, providing insights into agents' activities and intentions.
\end{enumerate}
This evolutionary transformation moves beyond basic spatial awareness to create a rich representation that captures goal assignment, task progress, coordination status, and behavioral dynamics—precisely the higher-level information agents need for effective coordination in referential communication tasks.

\section{Conclusion and Future Work}
\label{Conclusion}
In this paper, we proposed LERO, a novel framework that leverages large language models and evolutionary optimization to address two fundamental challenges in multi-agent reinforcement learning: credit assignment and partial observability. Our approach integrates LLM-generated hybrid reward functions that dynamically decompose team rewards into individual contributions, alongside observation enhancement functions that augment partial observations with inferred contextual information. These components are iteratively optimized through an evolutionary process that selects and refines high-performing candidates across MARL training cycles.

Our experimental results across multiple MPE scenarios and diverse MARL algorithms (MAPPO, VDN, and QMIX) demonstrate that LERO significantly outperforms baseline approaches. The framework achieves substantial performance improvements in complex multi-agent scenarios, with the full LERO implementation showing particularly impressive gains in environments requiring sophisticated coordination and information sharing. The ablation studies further confirm that while each component contributes meaningfully to performance enhancement, their integration within our evolutionary framework produces synergistic effects that exceed the sum of individual contributions.

In the future, we identify several promising directions: (1) scaling LERO to larger agent populations and more complex environments; (2) developing advanced evolutionary mechanisms with diversity preservation and adaptive mutation; (3) integrating LERO with cutting-edge MARL algorithms; and (4) reducing LLM inference computational costs. These improvements aim to enhance language model integration with reinforcement learning for solving complex multi-agent coordination problems.
%
%
%
\bibliographystyle{splncs04}
\bibliography{mybibliography}

\newpage
\appendix
\renewcommand{\thesection}{\Alph{section}} 
\section{Experimental Configuration Details}
\label{appendix A}
In our experiments, we conducted a thorough evaluation of the proposed framework across multiple environments and algorithms. 
To ensure reproducibility and provide a clear understanding of our experimental setup, we established standardized parameters for all evaluations. 

Tab. \ref{tab:exp_config} presents the complete configuration details, including environment parameters, LLM specifications, and training settings that were consistent across all experiments. 
For environment parameters, we adopted the default configuration from the MPE environment.
In terms of training settings, we carefully balanced training costs with performance outcomes to configure appropriate parameters that would demonstrate the effectiveness of our approach across diverse multi-agent scenarios without excessive computational requirements.
For our LLM component, we selected the o3-mini model from OpenAI, which offers strong reasoning capabilities while remaining computationally efficient. 

\begin{table}[htbp]
\centering
\caption{Experimental Configuration}
\begin{tabular}{p{6cm}p{3cm}}
\hline\hline
\multicolumn{2}{c}{\textbf{Environment Parameters}} \\
\hline
Number of agents ($N$) & $3^{spread} \& 2^{reference}$  \\
Number of landmarks ($M$) & 3 \\
Landmark coverage radius & 0.1 \\
Collision radius & 0.3 \\

\hline
\multicolumn{2}{c}{\textbf{Training Settings}} \\
\hline
Training steps & 30k \\
Evaluation episodes & 50 \\
Generated rewards per round & 3 \\
Evolution rounds & 4 \\
Population size & 2 \\
\hline
\multicolumn{2}{c}{\textbf{LLM Configuration}} \\
\hline
Model & o3-mini(OpenAI) \\
\hline\hline
\end{tabular}
\label{tab:exp_config}
\end{table}

\section{Prompt Design}
\label{appendix B}
In this section, we present our structured prompt design methodology for eliciting high-quality code generation from large language models. 
Each prompt template follows a three-part structure: a system prompt that defines the task parameters and expected output format, a user prompt that specifies the environmental context and requirements, and a feedback component that provides evaluation results to guide iterative refinement.
We apply this framework to generate two critical components for our multi-agent reinforcement learning system: reward functions that shape agent behavior and observation augmentation functions that enhance environmental representation.
\lstdefinestyle{plaintext}{
    basicstyle=\small\ttfamily,
    columns=flexible,
    breaklines=true,
    breakatwhitespace=true,
    frame=single,
    framesep=2mm,
    numbers=none,        
    showstringspaces=false,
    tabsize=2,
    captionpos=b
}
\begin{lstlisting}[style=plaintext, caption={Reward Function Prompt}]
<System>
You are a reward engineer designing reward functions for multi-agent reinforcement learning (MARL) tasks.
Your objective is to create two types of reward functions: a global reward and a agent reward.
Global Reward: Focus on maximizing the overall task performance for all agents.
Agent Reward: Encourage each agent to act optimally for its specific role.
There is a hybrid Reward combining the global and agent rewards to help agents learn effectively.
Based on the task described, an example of local and global reward functions might be: {task_reward_signature_string}.
Please strictly follow the above format to generate Python code snippets.

<User>
Write a reward function for the following task: 
{task_description}
The observation of the environment scenario:
{scenario_obs_code}
Ensure the reward functions uses only the available attributes and variables from the environment.
The code output should be formatted as a python code string: "```python ... ```".

<Feedback>
We trained the RL policy using provided reward function codes and tracked the values of the individual components in the reward function as well as global policy metrics.
{eval_result}
Please carefully analyze the current reward functions and policy feedback, provide a better agent reward function and a better global reward function.
Below is an illustrative example of the expected function format:
```python
{function_signature}
```
\end{lstlisting}
\begin{lstlisting}[style=plaintext, caption={Observation enhancement Function Prompt}]
<System>
You are a global state prediction engineer focusing on creating a **global information generation function** for a multi-agent reinforcement learning (MARL) task.  
Your job is to design a function (`stateEnhance`) that predicts **supplementary global information** based on the local observations (`obs`) of all agents.  
This global information will later be manually combined with the original `obs`, so your function does not need to process or modify the input `obs` directly. 
The function must:  
1. Extract **task-relevant global information** based on the input `obs`, such as:  
{global_information_advice}
2. Return a separate **global information tensor**.  
3. Use **efficient PyTorch tensor operations** (e.g., matrix computations) without relying on `for` loops for scalability. 
4. Be careful at the shape of variables in computation. Ensure all tensors are on the same device.  

Input and Output Details:  
- Input: A three-dimensional PyTorch tensor `obs` with the shape `(batch_size, num_agent, obs_length)`.  
  - `obs` contains the local observations of all agents, where each agent's observation is a vector of length `obs_length`.  
- Output: A three-dimensional PyTorch tensor with the shape `(batch_size, num_agent, global_info_length)`.  
  - `global_info_length` is a fixed-size vector representing the supplementary global information for each agent.  

Your function must focus on **extracting global information only** and ensure that the output information is **compact, task-specific, and meaningful**. 
Avoid including redundant or overly detailed data that could lead to unnecessary computational overhead.  
Focus on reducing dimensionality while preserving essential global information for reinforcement learning.
Below is an illustrative example of the expected function format:
```python
{function_signature}
```

<User>
The RL task description:
{task_description}
Original state description:
{scenario_state_code}
The code output should be formatted as a python code string: "```python ... ```".
Your task is to design a comprehensive state representation based on the local observations (obs) to improve reinforcement learning performance.
Ensure the function is executable and ready for integration into the RL environment.

<Feedback>
We trained the RL policy using provided state enhanced function codes and tracked the values of the individual components as well as global policy metrics.
{eval_result}
Please carefully analyze the current state enhanced function and policy feedback, provide a better state enhanced function.
Below is an illustrative example of the expected function format:
```python
{function_signature}
```
\end{lstlisting}

\section{Additional Experimental Results}
\label{appendix C}
To further investigate the impact of evolutionary optimization on our framework components, we conducted a comprehensive comparison between the evolutionary and non-evolutionary variants across different algorithms and environments. 
Tab.\ref{tab:evolutionary comparison} presents the performance analysis of our proposed approaches: LERO (LLM-driven Evolutionary framework with hybrid Rewards and enhanced Observation), LER (LLM-driven Evolutionary framework with hybrid Rewards), and LEO (LLM-driven Evolutionary framework with enhanced Observation), alongside their non-evolutionary counterparts LRO, LR, and LO respectively. 
\begin{table}[htbp]
    \centering
    \caption{Performance Analysis of Evolutionary Optimization}
    \begin{tabular}{cccccccc}
        \hline
        \multirow{2}{*}{Framework} & \multicolumn{3}{c}{Simple spread} & \multicolumn{3}{c}{Simple reference} \\
        \cline{2-7}
        & MAPPO & VDN & QMIX & MAPPO & VDN & QMIX \\
        \hline
        Baseline & 24.0\% & 38.0\% & 24.0\% & 23.0\% & 23.0\% & 29.0\% \\
        \rowcolor{gray!20}
        LR & 32.0\% & 33.3\% & 32.6\% & 24.0\% & 31.1\% & 29.0\% \\
        LER & 32.0\% & 36.6\% & 34.6\% & 24.0\% & 38.0\% & 31.1\% \\
        \rowcolor{gray!20}
        LO & 29.3\% & 38.0\% & 34.0\% & 23.0\% & 28.1\% & 37.1\% \\
        LEO & 74.7\% & 58.0\% & 34.0\% & 36.0\% & 33.0\% & 40.1\% \\
        \rowcolor{gray!20}
        LRO & 31.3\% & 38.0\% & 34.6\% & 23.0\% & 41.0\% & 37.1\% \\
        LERO & 74.7\% & 58.0\% & 48.0\% & 36.0\% & 83.0\% & 40.1\% \\
        \hline
    \end{tabular}
    \label{tab:evolutionary comparison}
\end{table}

The results demonstrate substantial performance gains from LERO framework, particularly in complex scenarios, with the full LERO framework achieving remarkable improvements of up to 83.0\% in the Simple reference environment with VDN, compared to the baseline performance of 23.0\%.

\section{Analysis of LLM generated functions}
\label{appendix D}
This appendix presents the complete outputs generated by LLM for both reward functions and observation augmentation functions. 
The following sections showcase four code snippets: (1) best hybrid reward functions without evolutionary(Lis.\ref{reward_no_evolutionary}), (2) best hybrid reward functions with evolutionary(Lis.\ref{reward_evolutionary}, (3) best observation enhancement function without evolutionary(Lis.\ref{obs_no_evolutionary}), and (4) best observation enhancement function with evolutionary(Lis.\ref{obs_evolutionary}). 
All Python code snippets were generated under the Simple Reference scenario using the Value Decomposition Networks (VDN) algorithm, which demonstrated the best overall performance in our framework evaluations.
\begin{lstlisting}[language=Python, caption={best hybrid reward function without evolutionary}, label={reward_no_evolutionary}]
def agent_reward(self, agent, world):
    """
    Calculates the individual reward for an agent based on how well it guides its
    partner (goal_a) to reach the assigned landmark (goal_b).

    The reward is computed as the negative Euclidean distance between the partner
    agent's position and the target landmark's position. A bonus is given if the partner
    reaches the landmark (i.e., the distance is less than 0.1).

    Args:
        agent: The agent for which the reward is being calculated.
        world: The environment world containing agents and landmarks.

    Returns:
        A scalar reward value.
    """
    # Ensure the agent has valid goal assignments
    if agent.goal_a is None or agent.goal_b is None:
        return 0.0

    # Compute the Euclidean distance between the partner agent and the target landmark
    dist = np.linalg.norm(agent.goal_a.state.p_pos - agent.goal_b.state.p_pos)
    
    # Base reward: minimizing the distance is beneficial
    reward = -dist

    # Provide a bonus if the partner reaches the target landmark
    if dist < 0.1:
        reward += 10.0

    return reward


def global_reward(self, world):
    """
    Calculates the global reward for the environment by aggregating the performance
    of all agents in guiding their respective partners towards their targets.

    This reward is the sum of all individual guidance outcomes. Each pair (agent, goal)
    contributes negatively by the distance between the partner and the target, with an added
    bonus if the target is reached (distance < 0.1).

    Args:
        world: The environment world containing agents and landmarks.

    Returns:
        A scalar total reward value for the entire environment.
    """
    total_reward = 0.0
    bonus = 10.0

    for agent in world.agents:
        if agent.goal_a is None or agent.goal_b is None:
            continue

        # Compute the Euclidean distance for the current guidance pair
        dist = np.linalg.norm(agent.goal_a.state.p_pos - agent.goal_b.state.p_pos)
        total_reward += -dist

        # Add bonus if the target landmark is reached
        if dist < 0.1:
            total_reward += bonus

    return total_reward
\end{lstlisting}
\begin{lstlisting}[language=Python, caption={best hybrid reward function with evolutionary}, label={reward_evolutionary}]
def agent_reward(self, agent, world):
    """
    Computes an improved individual reward for the guiding agent in the simple_reference task.
    This reward is based on multiple factors:
      1. Minimizing the distance of the assigned partner (goal_a) to the target landmark (goal_b).
      2. Encouraging the guiding agent to stay close to its partner to better coordinate the task.
      3. Providing a bonus for stability when the partner is moving slowly near the target.
      4. Incorporating a shaping term to increase the incentive when the distance is significantly reduced.
    
    Args:
        agent: The current guiding agent.
        world: The environment containing all agents and landmarks.
    
    Returns:
        rew (float): The scalar reward for the guiding agent.
    """
    # Check if the agent goals are assigned
    if (agent.goal_a is None) or (agent.goal_b is None):
        return 0.0

    # Calculate the distance from the partner (goal_a) to the target (goal_b)
    d_target = ((agent.goal_a.state.p_pos - agent.goal_b.state.p_pos) ** 2).sum() ** 0.5

    # The basic shaping reward: a high penalty that is eased as the agent approaches the target
    rew = -2.0 * d_target

    # Enhanced shaping: if partner is within near-target range, give a high bonus
    if d_target < 0.1:
        rew += 15.0
    elif d_target < 0.2:
        rew += 8.0
    elif d_target < 0.3:
        rew += 4.0

    # Bonus for maintaining close proactive communication: keep the guiding agent near its partner
    d_partner = ((agent.state.p_pos - agent.goal_a.state.p_pos) ** 2).sum() ** 0.5
    if d_partner < 0.3:
        rew += 3.0
    elif d_partner < 0.5:
        rew += 1.5
    else:
        rew -= 2.0  # Penalize if too separated

    # Stability bonus: if the partner is nearly at the target and moving slowly, reward further
    partner_speed = ((agent.goal_a.state.p_vel) ** 2).sum() ** 0.5
    if d_target < 0.1 and partner_speed < 0.05:
        rew += 5.0

    return rew


def global_reward(self, world):
    """
    Computes an improved global reward for the collaborative task in simple_reference.
    The reward encourages all guiding pairs to bring their partner agents close to the correct
    landmarks and is constructed using:
      1. A cumulative penalty based on each partner agent's distance to its target.
      2. Shaping bonuses for pairs that are close enough.
      3. A significant collective bonus if all pairs achieve the near-target criteria.
    
    Args:
        world: The environment containing all agents and landmarks.
    
    Returns:
        rew (float): The global collaborative reward.
    """
    total_rew = 0.0
    success_count = 0

    for agent in world.agents:
        # Skip agents without goals assigned
        if (agent.goal_a is None) or (agent.goal_b is None):
            continue

        # Distance from partner (goal_a) to target (goal_b)
        d_target = ((agent.goal_a.state.p_pos - agent.goal_b.state.p_pos) ** 2).sum() ** 0.5

        # Base penalty reward for distance
        total_rew += -2.0 * d_target

        # Shaping bonuses based on nearness to the target landmark
        if d_target < 0.1:
            total_rew += 15.0
            success_count += 1
        elif d_target < 0.2:
            total_rew += 8.0
        elif d_target < 0.3:
            total_rew += 4.0

    # Provide an extra collective bonus when all pairs reach near the target (aiming to raise coverage rate)
    if success_count == len(world.agents) and len(world.agents) > 0:
        total_rew += 20.0

    return total_rew
\end{lstlisting}

\begin{lstlisting}[language=Python, caption={best observation enhancement function without evolutionary}, label={obs_no_evolutionary}]
def stateEnhance(obs):
    """
    Enhances agent observations by extracting supplementary global information.
    
    Specifically, for each agent this function computes:
      - Euclidean distances to each of the 3 landmarks.
      - An estimated distance to the partner agent based on the landmark-relative vectors.
      
    Args:
        obs (torch.Tensor): A tensor of shape (batch_size, num_agent, obs_length)
            representing local observations. Expected obs_length is 21 with the following structure:
              [self_vel (2-dim), all_landmark_rel_positions (6-dim), 
               goal_color (3-dim), communication (10-dim)]
    
    Returns:
        torch.Tensor: A tensor of shape (batch_size, num_agent, global_info_length) where
                      global_info_length = 4. The 4 features per agent are:
                        - Three distances to landmarks.
                        - Estimated distance to partner agent.
                        
    Notes:
      - This is designed for a 2-agent scenario as in the PettingZoo MPE simple_reference task.
      - All tensor operations are vectorized (using PyTorch matrix operations) and without Python loops.
      - The function assumes that obs is on the correct device.
    """
    # Get the device and tensor shape
    device = obs.device
    batch_size, num_agent, obs_length = obs.shape  # obs_length should be 21

    # Extract landmark relative positions.
    # In the provided observation, indices 2 to 7 correspond to the concatenated relative positions of 3 landmarks.
    # Reshape from (batch_size, num_agent, 6) to (batch_size, num_agent, 3, 2).
    landmark_rel = obs[:, :, 2:8].reshape(batch_size, num_agent, 3, 2)
    
    # Compute the Euclidean distances from each agent to each landmark.
    # Resulting shape: (batch_size, num_agent, 3)
    landmark_dists = torch.norm(landmark_rel, dim=-1)
    
    # Estimate the distance between agents (partner distance).
    # For the simple_reference task, there are exactly 2 agents.
    # We use the fact that each agent's observation provides landmark-relative vectors,
    # and by comparing the corresponding vectors from both agents we can estimate the
    # relative displacement between them.
    if num_agent == 2:
        # Compute the difference in landmark-relative vectors between the two agents.
        # landmark_rel[:, 0, :, :] is for agent 0 and landmark_rel[:, 1, :, :] is for agent 1.
        # Theoretically, since for a landmark: landmark_rel = landmark_pos - agent_pos,
        # the difference landmark_rel_0 - landmark_rel_1 = (landmark_pos - pos0) - (landmark_pos - pos1)
        # = pos1 - pos0. Averaging over multiple landmarks can help reduce noise.
        diff = landmark_rel[:, 0, :, :] - landmark_rel[:, 1, :, :]  # shape: (batch_size, 3, 2)
        avg_diff = diff.mean(dim=1)  # shape: (batch_size, 2)
        # Compute the partner distance as the norm of the average difference.
        pdist = torch.norm(avg_diff, dim=-1, keepdim=True)  # shape: (batch_size, 1)
        # Duplicate the partner distance for each agent to form a per-agent scalar feature.
        partner_dist = pdist.unsqueeze(1).repeat(1, num_agent, 1)  # shape: (batch_size, num_agent, 1)
    else:
        # For tasks with more than 2 agents, one might compute a pairwise agent distance matrix
        # and then select or aggregate relevant distances. For now, we set partner_dist to zero.
        partner_dist = torch.zeros(batch_size, num_agent, 1, device=device)

    # Concatenate the computed global features:
    # - Landmark distances (3-dimensional per agent)
    # - Partner distance (1-dimensional per agent)
    # This yields a global information tensor of shape: (batch_size, num_agent, 4)
    global_info = torch.cat([landmark_dists, partner_dist], dim=-1)

    return global_info
\end{lstlisting}

\begin{lstlisting}[language=Python, caption={best observation enhancement function with evolutionary}, label={obs_evolutionary}]
def stateEnhance(obs):
    """
    An improved global state enhancement function for the PettingZoo MPE simple_reference task.
    
    In this revision, we augment the global information by directly comparing the distance
    (and angle) of the self agent and its partner relative to their assigned landmarks. In
    addition, we include helpful motion features (velocity norms) and a computed communication
    channel magnitude for the partner.
    
    For each agent, the function extracts 9 feature components:
      1. Self assigned landmark distance.
      2. Self assigned landmark angle.
      3. Coverage margin: max(0, 0.1 - self landmark distance).
      4. Partner assigned landmark distance.
      5. Partner assigned landmark angle.
      6. Relative assigned distance difference: (self distance - partner distance).
      7. Self velocity norm.
      8. Partner velocity norm.
      9. Partner communication channel norm.
    
    Assumptions:
      - Input `obs` tensor shape: (batch_size, num_agent, obs_length), with obs_length = 21.
      - Observation indices:
            [0:2]   : self_vel (2-dim)
            [2:8]   : relative positions of 3 landmarks (3 x 2 = 6-dim)
            [8:11]  : goal_color (3-dim)
            [11:21] : communication (10-dim)
      - Landmark color references (fixed):
            red   : [0.75, 0.25, 0.25]
            green : [0.25, 0.75, 0.25]
            blue  : [0.25, 0.25, 0.75]
      - The assigned landmark for an agent is inferred by matching its goal_color to the fixed colors.
      - A landmark is considered "covered" if the agent-to-landmark distance is less than 0.1.
      - Designed primarily for the 2-agent case. For other configurations, partner features are set to zero.

    Returns:
      A tensor of shape (batch_size, num_agent, 9) containing the augmented global information.
    """
    device = obs.device
    batch_size, num_agent, obs_length = obs.shape  # obs_length is expected to be 21

    # Fixed landmark colors (3 landmarks) on the proper device and with the same dtype as obs.
    fixed_landmark_colors = torch.tensor(
        [[0.75, 0.25, 0.25],
         [0.25, 0.75, 0.25],
         [0.25, 0.25, 0.75]],
        device=device,
        dtype=obs.dtype
    )  # shape: (3, 3)
    
    # ---------------------------
    # 1. Self Agent Features
    # ---------------------------
    # Extract self-relative landmark positions from indices 2 to 8 and reshape to (batch, agent, 3, 2)
    self_landmark_rel = obs[:, :, 2:8].reshape(batch_size, num_agent, 3, 2)
    # Euclidean distances for each landmark: shape (batch, agent, 3)
    self_landmark_dists = torch.norm(self_landmark_rel, dim=-1)
    # Compute landmark angles using arctan2: shape (batch, agent, 3)
    self_landmark_angles = torch.atan2(self_landmark_rel[..., 1], self_landmark_rel[..., 0])
    
    # Extract self goal color from indices 8 to 11: shape (batch, agent, 3)
    self_goal_color = obs[:, :, 8:11]
    # Compute squared differences to fixed landmark colors and sum along color channel.
    # Expand dimensions to (batch, agent, 1, 3) and (1, 1, 3, 3) respectively.
    color_diff = (self_goal_color.unsqueeze(2) - fixed_landmark_colors.unsqueeze(0).unsqueeze(0)) ** 2
    color_error = torch.sum(color_diff, dim=-1)  # shape (batch, agent, 3)
    # Determine the assigned landmark index by taking argmin of color error.
    self_assigned_idx = torch.argmin(color_error, dim=-1)  # shape (batch, agent)
    # Gather self assigned landmark distance and angle.
    self_assigned_dist = torch.gather(self_landmark_dists, dim=2, index=self_assigned_idx.unsqueeze(-1))  # (batch, agent, 1)
    self_assigned_angle = torch.gather(self_landmark_angles, dim=2, index=self_assigned_idx.unsqueeze(-1))  # (batch, agent, 1)
    # Compute coverage margin: how much within threshold (0.1) the agent is from the landmark.
    coverage_margin = torch.clamp(0.1 - self_assigned_dist, min=0)  # (batch, agent, 1)
    
    # ---------------------------
    # 2. Partner Agent Features (Assuming 2-agent scenario)
    # ---------------------------
    if num_agent == 2:
        # Use a roll to get partner's observation.
        partner_obs = obs.roll(shifts=1, dims=1)
    
        # Partner relative landmark positions.
        partner_landmark_rel = partner_obs[:, :, 2:8].reshape(batch_size, num_agent, 3, 2)
        partner_landmark_dists = torch.norm(partner_landmark_rel, dim=-1)
        partner_landmark_angles = torch.atan2(partner_landmark_rel[..., 1], partner_landmark_rel[..., 0])
        
        # Partner goal color.
        partner_goal_color = partner_obs[:, :, 8:11]
        partner_color_diff = (partner_goal_color.unsqueeze(2) - fixed_landmark_colors.unsqueeze(0).unsqueeze(0)) ** 2
        partner_color_error = torch.sum(partner_color_diff, dim=-1)  # (batch, agent, 3)
        partner_assigned_idx = torch.argmin(partner_color_error, dim=-1)  # (batch, agent)
        partner_assigned_dist = torch.gather(partner_landmark_dists, dim=2, index=partner_assigned_idx.unsqueeze(-1))  # (batch, agent, 1)
        partner_assigned_angle = torch.gather(partner_landmark_angles, dim=2, index=partner_assigned_idx.unsqueeze(-1))  # (batch, agent, 1)
        
        # Compute partner velocity norm from indices 0 to 2.
        partner_vel = partner_obs[:, :, 0:2]
        partner_vel_norm = torch.norm(partner_vel, dim=-1, keepdim=True)  # (batch, agent, 1)
        
        # Compute partner communication channel norm from indices 11 to 21.
        partner_comm = partner_obs[:, :, 11:21]
        partner_comm_norm = torch.norm(partner_comm, dim=-1, keepdim=True)  # (batch, agent, 1)
    else:
        partner_assigned_dist = torch.zeros(batch_size, num_agent, 1, device=device, dtype=obs.dtype)
        partner_assigned_angle = torch.zeros(batch_size, num_agent, 1, device=device, dtype=obs.dtype)
        partner_vel_norm = torch.zeros(batch_size, num_agent, 1, device=device, dtype=obs.dtype)
        partner_comm_norm = torch.zeros(batch_size, num_agent, 1, device=device, dtype=obs.dtype)
    
    # ---------------------------
    # 3. Additional Self Features
    # ---------------------------
    # Self velocity norm from indices 0 to 2.
    self_vel = obs[:, :, 0:2]
    self_vel_norm = torch.norm(self_vel, dim=-1, keepdim=True)  # (batch, agent, 1)

    # Compute relative difference of assigned distances (self - partner)
    rel_dist_diff = self_assigned_dist - partner_assigned_dist  # (batch, agent, 1)
    
    # ---------------------------
    # 4. Concatenate Global Features
    # ---------------------------
    # Final Global Info per agent: 9 dimensions.
    # [self_assigned_dist, self_assigned_angle, coverage_margin,
    #  partner_assigned_dist, partner_assigned_angle, rel_dist_diff,
    #  self_vel_norm, partner_vel_norm, partner_comm_norm]
    global_info = torch.cat([
        self_assigned_dist,     # (batch, agent, 1)
        self_assigned_angle,    # (batch, agent, 1)
        coverage_margin,        # (batch, agent, 1)
        partner_assigned_dist,  # (batch, agent, 1)
        partner_assigned_angle, # (batch, agent, 1)
        rel_dist_diff,          # (batch, agent, 1)
        self_vel_norm,          # (batch, agent, 1)
        partner_vel_norm,       # (batch, agent, 1)
        partner_comm_norm       # (batch, agent, 1)
    ], dim=-1)
    
    return global_info
\end{lstlisting}

\end{document}